%% file: main.tex
\documentclass[10pt,twocolumn,letterpaper]{article}

\usepackage{multirow}
 \usepackage[pagenumbers]{cvpr} %

\input{preamble}

\usepackage{booktabs}
\usepackage{caption}
\usepackage{cuted}
\usepackage{dblfloatfix}
\usepackage{amsthm}

\definecolor{cvprblue}{rgb}{0.21,0.49,0.74}
\usepackage[pagebackref,breaklinks,colorlinks,citecolor=cvprblue]{hyperref}

\include{new_commands}
\renewcommand{\paragraph}[1]{\vspace{0.5em}\noindent\textbf{#1}}
\newcommand{\KL}[2]{\operatorname{KL}\big(#1\,\|\,#2\big)}
\newcommand{\E}{\mathbb{E}}

\DeclareMathOperator*{\argmin}{arg\,min}

\usepackage{bbm}

\def\paperID{13147} %
\def\confName{CVPR}
\def\confYear{2024}

\title{%
Interpretable Measures of Conceptual Similarity by\\  Complexity-Constrained Descriptive Auto-Encoding
}

\author{Alessandro Achille$^*$ \quad Greg Ver Steeg$^*$\quad Tian Yu Liu \quad Matthew Trager \\  Carson Klingenberg \quad Stefano Soatto \\
{\tt\small } AWS AI Labs%
}

\begin{document}
\maketitle

\begin{abstract}
Quantifying the degree of similarity between images is a key copyright issue for image-based machine learning. In legal doctrine however, determining the degree of similarity between works requires subjective analysis, and fact-finders (judges and juries) can demonstrate considerable variability in these subjective judgement calls. Images that are structurally similar can be deemed dissimilar, whereas images of completely different scenes can be deemed  similar enough to support a claim of copying. We seek to define and compute a notion of ``conceptual similarity'' among images that captures high-level relations even among images that do not share repeated elements or visually similar components. The idea is to use a base multi-modal model to generate ``explanations'' (captions) of visual data at increasing levels of complexity. Then, similarity can be measured by the length of the caption needed to discriminate between the two images: Two highly dissimilar images can be discriminated early in their description, whereas conceptually dissimilar ones will need more detail to be distinguished. We operationalize this definition and show that it correlates with subjective (averaged human evaluation) assessment, and beats existing baselines on both image-to-image and text-to-text similarity benchmarks. Beyond just providing a number, our method also offers interpretability by pointing to the specific level of granularity of the description where the source data are differentiated.
\end{abstract}

\vspace{-0.5em}

\begin{figure*}[!ht]
    \centering
    \includegraphics[width=.9\linewidth]{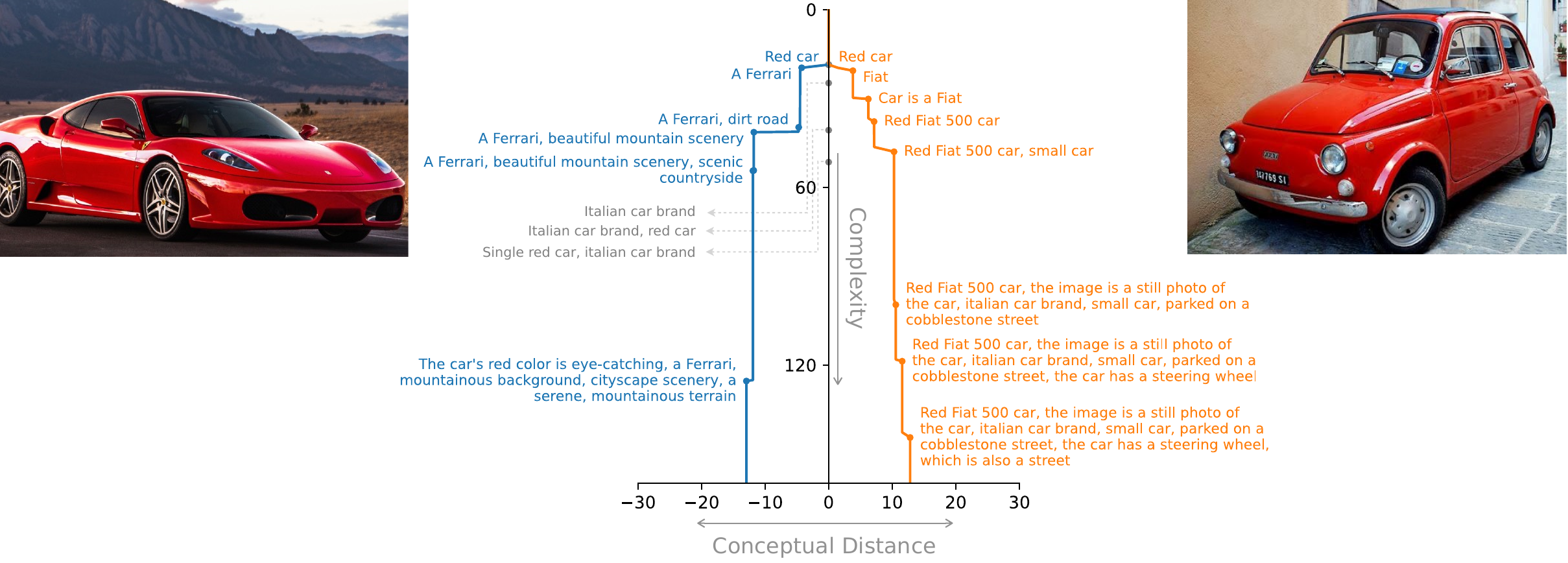}
    \caption{If we describe each image at increasing levels of complexity (blue and orange text), short descriptions apply equally well to both, as measured by their likelihood. However, as the complexity level of the description increases, a gap emerges between the likelihood under the best common description (grey) and the likelihood under the best individual descriptions (blue and orange).   For instance, at $C=36$ the best individual descriptions are ``Red Fiat 500 car'' and ``A Ferrari'' whereas the best common description is ``Italian car brand'' which is not as descriptive.  The gap traces two asymmetric curves that measure the conceptual difference between the images at each level of complexity.  A single number can be obtained by measuring the area under between curves.  
    \label{fig:splash-figure1}}
\end{figure*}

\section{Introduction}
\label{sec:introduction}

\begin{figure}[t]
    \centering

    \includegraphics[width=0.48\linewidth]{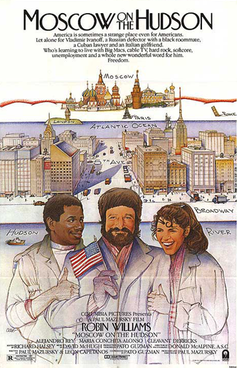}
    \includegraphics[width=0.48\linewidth]{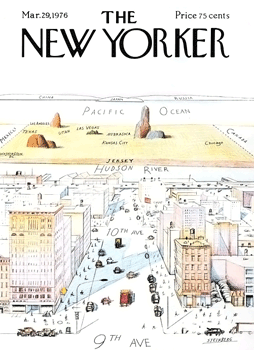}

    \caption{These two images have similar art styles, theme, and subject matter. On the other hand, it is difficult to identify specific visual elements that appear in both images. These two images were found to be ``substantially similar'' \cite{steinberg_v_columbia} based on the arrangement of \textit{similar} features in a \textit{similar} way. How can we measure how similar these images are? 
    }
    \label{fig:WOF}
    \vspace{-1em}
\end{figure}

Consider the two images in Fig.~\ref{fig:splash-figure1}. One could say they are similar: both portray small red Italian cars. %
Another could say they are different: One is a sports car in an open space, the other a tiny city car in an alley. Is there an objective way of measuring the similarity among images? %
In some cases, similarity judgments can be influenced by shared concepts: in Fig.~\ref{fig:WOF}, two images share compelling stylistic and conceptual similarity, but it is difficult to identify specific visual elements they share. Yet the two were found to be legally ``substantially similar'' \cite{steinberg_v_columbia}. Can we define an objective notion of ``conceptual similarity'' among data?

There have been many attempts at defining an objective notion of similarity based on the number of bits of information that the two samples share \cite{li2008introduction,viola1997alignment,cover1999elements}, but fundamentally they do not capture concepts (see Appendix), which are human constructs. Since humans must play a role in defining similarity among concepts, one way to achieve objectivity is by averaging a large number of human assessments. This is what large-scale neural network models do \cite{radford2021learning}. However, contrastive-trained models measure similarity based on how easily the data can be distinguished, and any two non-identical samples can be distinguished by random features of no conceptual relevance \cite{johnsonlindenstrauss}. 

Rather than focusing on finding the features that best distinguish the two images, we focus on finding the ones that best {\em describe} them. Then, we can measure semantic similarity based on how well descriptions of one fit the other. If two samples are similar, a short description should apply equally well to both. If the samples are very different, the description of one will be a poor fit for the other. The more complex the description needed to distinguish the images, the higher their conceptual similarity. The key idea is to focus on optimally describing individual samples, rather than adversarially discriminating them. 

Specifically, referring to Fig.~\ref{fig:splash-figure1}, we generate multiple descriptions of each sample in increasing level of complexity, sorted by their coding length. Then, we measure the difference of the likelihood of each image conditioned on either descriptions as a function of complexity. That traces a curve that measures distance as a function of complexity. Any two images can be distinguished by their description {\em at some point} (the more you look, the more {\em differences} you see). Therefore, similarity should always be relative to a description length. However, when a single number is needed for comparison, we show that the AUC of the distance function is well aligned with human similarity assessments. 

Our proposed method, which we call Complexity-Constrained Descriptive Autoencoding, or CC:DAE, is rooted in the idea of the Kolmogorov Structure Function, which was introduced to differentiate semantic ``information'' from structureless ``noise.'' But while Kolmogorov used \textit{programs} to describe the data, and program length as their complexity, we use natural language descriptions and their coding length. 
One may wonder whether our notion of conceptual distance is canonical in any sense, since it hinges on arbitrary choices, among which the use of language, or even a particular trained language model. Indeed, as we prove formally, no notion of conceptual similarity can be canonical: Using results from \cite{gacs2001algorithmic}, we show that Kolmogorov's choice itself does not convey any information about the semantics of data and more generally that no non-trivial notion of common conceptual information can be defined without imposing strong restrictions on the class of representations and encoder/decoder used. The unavoidable need to perform subjective analysis is recognized by copyright legal doctrines, which provide generalized guidelines for analyzing similarity but leave the ultimate determination to be made on a case-by-case basis. As a result, despite efforts to codify similarity into guidelines, one is left with the impression that there are as many notions of similarity as there are judges or juries. One of the advantages of large-scale pre-trained models is that they aggregate content from vast and diverse corpora and, unlike their human sources, their biases can be measured, monitored, and calibrated. 

Our method applies to general data, including any combination of text and image and any choice of data-to-text encoder ({\em e.g.,} captioning models) and a text-to-data generative model ({\em e.g.,} multi-modal autoregressive models or diffusion models). It can also be used to measure similarity of data across different modalities ({\em e.g.,} between text and images).

To evaluate the alignment between our notion of conceptual similarity and human assessments, we use established human-annotated similarity benchmarks, and obtain state-of-the-art results on the STS sentence similarity benchmarks beating all methods that have not explicitly been trained on human annotated similarity scores. We also surpass CLIP on the CxC-SIS image similarity benchmark.

\section{Related Work}
Using machine learning to represent conceptual information is a long-running yet elusive goal in machine learning. 
For instance, the ``concept bottleneck'' \cite{koh2020concept} epitomizes a line of work that generate restricted representations based on human specified concepts. Our approach is more general: we do not specify specific concepts, but only demand natural language descriptions with constrained complexity. 
Lossy compression, for example through capacity-contrained variational auto-encoders \cite{higgins2016beta,burgess2018understanding}, is a common approach to try to distinguish semantic information in images 
\cite{gregor2016towards},
but the success of these methods is largely due to inductive priors coming from the architecture. In general, there is no reason to expect compression alone to align with human intuition of conceptual information, as we demonstrate in \cref{sec:motivations}. 
Another line of work referring to ``visual concept learning'' defines concepts in terms of a set of generalizable properties that then can be used in conjunction with neuro-symbolic programs to answer novel queries \cite{falcon}, or in terms of embeddings that can be used to generate images \cite{liu2023unsupervised}.

Recent work on contrastive vision-language models \cite{radford2021learning} has demonstrated that these models are surprisingly weak at understanding semantic relationships~\cite{yuksekgonul2022,conwell2022testing}, with only modest improvements from conditional diffusion models~\cite{kong2023interpretable}.
Generating natural language descriptions of images as representations was also considered by \cite{bucher2019semantic}, who finds the resulting description valuable for a visual question answering task and more useful than the original images for a retrieval task. 

Embedding generated by auto-regressive models fine-tuned on human annotations are also frequently used to quantify similarity \cite{devlin2018bert,ni2021sentence,reimers2019sentence,zhang2020unsupervised,gao2021simcse,jiang2022promptbert}. Fewer works define similarity metrics using pre-trained auto-regressive models without supervised fine-tuning: \cite{jiang2023scaling} uses prompts to condition model outputs; \cite{muennighoff2022sgpt} evaluates similarity of two samples via the conditional probability of one given the other. 
Most similar to our work, \cite{liu2023meaning} defines semantic similarity as the divergence between conditional distributions over future trajectories. Interestingly, \cite{liu2023meaning} can be considered a special case of our method (see Appendix), which further benefits from improved interpretability.

\section{CC:DAE}
\label{sec:concept}

In this section we formalize the critical components of our method, including the optimal description of the data using complexity-constrained textual expressions, and the computation of the fit of the description, from which the conceptual distance is measured.

\subsection{Optimal Description}

Let $x \in X$ be a sample and let $H = \{h_1, h_2, \ldots\}$ be space of hypotheses. We think of each $h \in H$ as a possible description of the sample, for example a natural language sentence. Associated with $H$ and $X$ we consider a \textit{decoder} $p(x|h)$ which computes the conditional likelihood of the input $x$ given the description $h$. We refer to the negative conditional log-likelihood $\ell(x|h) = -\log p(x|h)$ as the reconstruction loss of $x$ under hypothesis $h$. A key quantity for us is the \textit{complexity}, or \textit{coding length}, $\ell(h)$ of an hypothesis $h$. By Kraft's inequality,  we can express the number of bits needed to encode an hypothesis $h$ as a negative log-likelihood $\ell(h) = -\log \p(h)$ for some distribution $\p(h)$. 
Conversely, for a given $p(h)$, a lossless compression scheme whose code-lengths almost exactly match $-\log p(h)$ can be constructed with arithmetic coding.

\paragraph{Example.} Let $X$ be the space of all images and let $H$ be the set of all English sentences. The cost of encoding $h \in H$ can be computed as the negative log-likelihood $-\log \p(h)$ assigned to $h$ by a trained large language model (LLM). The log-likelihood $\log p(x|h)$ of an image $x$ given a sentence $h$ can be computed using a conditional image generation model, such as a diffusion model.

\paragraph{Optimal descriptions.} Suppose we want to describe a sample $x$, but we have an upper-bound $\ell(h) \leq C$ on the length of the description $h$ we can use. To find the best description $h^*_x(C)$ to use under this capacity constraint, we have to solve the constrained optimization problem:
\begin{align}
    \label{eq:discrete-description}
    h^*_x(C) = \argmin_{h \in H}& \ \ \ell(x|h) \\
    \text{s.t.}& \ \ -\log \p(h) \leq C, \nonumber
\end{align}
which aims at finding the hypothesis $h$ with $\ell(h) \leq C$ which minimizes the reconstruction error for $x$. This definition however has practical limitations: since the space $H$ is discrete, the optimization in \cref{eq:discrete-description} cannot be performed easily with gradient-based methods, leading to an expensive search over the (potentially infinite) hypothesis space $H$. 

\paragraph{Stochastic relaxation.} To simplify the problem, rather than considering the best \textit{single} hypothesis, we can search for a \textit{distribution} of hypotheses $q(h)$ describing the image. This corresponds to a stochastic relaxation of the problem where we find a low-complexity \textit{distribution} of hypotheses $q^*_x(h|C)$ that, on average, have good reconstruction loss:
\begin{align}
    \label{eq:stochastic-description}
    q^*_x(h|C) = \argmin_{q(h) \in \mathcal{P}(H)}& \ \ \E_{h \sim q(h)} [ \ell(x|h)] \\
    \text{s.t.}& \ \ \KL{q(h)}{\p(h)} \leq C, \nonumber
\end{align}
where the optimization is now over the space of distributions $q(h) \in \mathcal{P}(H)$.
When $q(h)$ is restricted to being a Dirac delta over a single hypothesis --- i.e., $q(h) = \delta_{h_0}(h)$ --- we exactly recover \cref{eq:discrete-description}.
A natural question is in which way the newly introduced term $\KL{q(h)}{p(h)}$ generalizes the description length $-\log \p(h)$ of \cref{eq:discrete-description}, aside from reducing to it in the extreme case where $q(h)$ is a Dirac delta. To answer, it can be shown that $\KL{q(h)}{p(h)}$ is the expected number of bits required to specify a fair sample of the distribution $q(h)$ to a receiver that knows $p(h)$ but not $q(h)$ \cite[Lemma 1.5]{harsha2007communication}. For this reason, we refer to \cref{eq:stochastic-description} as a stochastic relaxation of \cref{eq:discrete-description}. The advantage of this formulation is that the solution to the optimization problem can be efficiently approximated by sampling a large language model, as we will discuss later. This formulation also connects to other frameworks (see \cref{sec:motivations}).

\subsection{Conceptual Similarity}

As anticipated in \Cref{sec:introduction}, given two sample $x_1$ and $x_2$, we want to define a notion of \textit{conceptual similarity} by measuring how well optimal descriptions of $x_1$ also apply to $x_2$ --- and vice versa -- as the complexity of the description increases.
To formalize this notion, let $q_i(h|C)$ denote the family of distributions of optimal descriptions $h$ for the sample $x_i$ as $C$ varies.
We can define the function $\Delta_{2 \to 1}(C)$ 
\begin{align}
    \Delta_{2 \to 1}(C) := \E_{q_1}[ \ell(x_1|h)] - \E_{q_2}[\ell(x_1|h)].
\end{align}
which measures how well the description $q_2$ of $x_2$ describe $x_1$, compared to its optimal description $q_1$. Similarly, we can define $\Delta_{1 \to 2}(C)$ inverting the role of $x_1$ and $x_2$.

\paragraph{Conceptual Distance.} We define the \textit{conceptual distance} between $x_1$ and $x_2$ at complexity level $C$ as
\begin{align}
    \label{eq:conceptual-distance}
    d_{x_1,x_2}(C) =& \frac{1}{2} \Big(\Delta_{2 \to 1}(C) + \Delta_{1 \to 2}(C)\Big).
\end{align}
\Cref{fig:splash-figure1} provides a sample illustration of the typical behavior of this distance. For low values of $C$ (y-axis), the optimal descriptions of one sample --- e.g., ``a red car'' --- is an equally good description of both. In this case, both terms $\Delta_{i\to j}$ (blue and orange curves in the plot) will be zero. As the complexity $C$ increases, the optimal description of one sample --- e.g., ``a Ferrari'' --- is not an optimal description for the other and the distance becomes non-zero, and will keep increasing as the descriptions become more detailed and specific.

\paragraph{AUC.} Rather than being a single number, our definition of conceptual distance is a function of the complexity $C$ of the description. As discussed in \Cref{sec:motivations}, this is necessary to address some key issues in defining distances and we posit that human perception of similarity relates to how quickly this function grows. To capture this, when a single number is necessary to define a distance, we use the Area Under Curve of the graph, up to some vale $C_\text{max}$.

\paragraph{Asymmetric distances.} The distance $d_{x_1,x_2}(C)$ in \cref{eq:conceptual-distance} in written in terms of two {\em asymmetric distances}, the distance $\Delta_{1\to 2}$ of $x_1$ from $x_2$, and $\Delta_{2\to 1}$ from $x_2$ to $x_1$. While $d_{x_1,x_2}(C)$ provides an overall measure of the distance between samples, each individual asymmetric distances can be useful to measure directed containment relations, for instance to what extent a work is being subsumed by or supersedes another. 

\paragraph{Interpretation with intersection.}
We note that, rearranging the terms, \cref{eq:conceptual-distance} can also be rewritten as:
\begin{align}\label{eq:d}
    d_{x_1,x_2}(C) =&\, \E_{q_1}[\ell(x_1|h)] + \E_{q_2}[\ell(x_2|h)] \nonumber \\
    &\,- \E_{q_\cap}[\ell(x_1|h) + \ell(x_2|h)]
\end{align}
where $q_\cap(h|C) = \frac{1}{2} q_1(h|C) + \frac{1}{2} q_2(h|C)$. This compares the reconstruction loss obtained by using different optimal descriptions for $x_1$ and $x_2$ and the reconstruction loss obtained by encoding both $x_1$ and $x_2$ using the \textit{same} descriptions $h \sim q_\cap(h|C)$. Intuitively, if the samples are similar, a shared description equally described each image when compared to picking the optimal descriptions independently. However, as the complexity of $C$ increases, $q_1(h|C)$ and $q_2(h|C)$ will contain increasing details that describe one sample $x_1$ (respectively $x_2$) but not equally well the other. In that case $q_\cap$ becomes suboptimal and the distance grows accordingly.

\paragraph{Interpretability.} An advantage of our definition of conceptual distance is that the descriptions $h$ that have high probability under $q_\cap$ provide an interpretable explanation for why the two samples should be considered similar (e.g., gray captions in \Cref{fig:splash-figure1}). Conversely, looking at $h$ that are likely under $q_i$ but not $q_\cap$ provides an explanation of the unique information in $q_i$, justifying why $q_1$ and $q_2$ should be considered different.

\section{Motivation and Discussion}
\label{sec:motivations}

Defining a conceptual distance face three main challenges:
\begin{enumerate}[label = (\roman*)]
    \item \textbf{Randomness dominates.} Any pair of non-identical real images has a large number of small differences due to randomness which dominates over the few, but important, structural similarities;
    \item \textbf{Canonical importance of properties.} What properties are important for similarity is non-canonical;
    \item \textbf{Adversarial discrimination.} Any two images may be ``adversarially'' distinguished by simple and obvious properties (e.g., ``the color of the car is different'') but ease of \textit{discriminability} should not affect \textit{similarity} (two pictures of the same car model should be similar, even if the color is different).
\end{enumerate}
Corresponding to these challenges there have are three main classes of distance functions that address some of the problems, but fail to capture the others.

\paragraph{Contrastive learning.} CLIP-like contrastive learning models by their nature rely on discriminative features that are optimized to discriminate samples, not to describe concepts that humans evince from said samples, thus failing (iii). Since these representations are not trained to align with human conceptual representations, they are a poor fit to measure conceptual similarity as noted in recent works~\cite{yuksekgonul2022,conwell2022testing}. Indeed, we show in \cref{tab:cxc-benchmarks} that our method significantly outperforms CLIP in outputting human-aligned similarity scores for images. The method also fails (ii) since the model learns semantically relevant properties either through semi-supervised image-text or enforced through hard-coded strong data augmentation.

\paragraph{Information theoretic  distances}, such as the Normalized Compression Distance (NCD), move away from discriminability and instead define a canonical notion of similarity between samples as the ratio between the amount of shared (algorithmic) information and the total information. However, reflecting (i), unique randomness in the data accounts for most of the information in high dimensional samples. Hence, even when samples are perceptually similar, random information will dominate the distance and similarity of key semantic information is lost (see Appendix).

\paragraph{Structure function.} Defining a distance for high-dimensional data requires separating ``structural'' information of the sample from the information due to randomness and noise, raising the question of how to canonically define what constitutes ``structure'' and what ``noise''. 
Kolmogorov proposed that describing structural information should reduce the reconstruction loss of a sample more quickly than describing random details which are intrinsically incompressible.
This is captured by the Structure Function $\beta_x(C)$, defined as
\begin{align}
    \label{eq:structure}
    \beta_x(C) = \min_{h \in H}& \ \ -\log p_h(x) \\
    \text{s.t.}& \ \ |h| \leq C, \nonumber
\end{align}
which measures the reconstruction error when compressing $x$ using a program $h \in H$ implementing a computable probability distribution $p_h(x)$, as the length $|h|$ of the program varies. 
Since the optimal descriptions have to reduce the reconstruction loss as quickly as possible, for low values of $C$ they can only contain structural information and not random information. This reasoning can be formalized \cite{vitanyi} and allows to define a notion of \textit{algorithmic minimal sufficient statistic} that captures structural but not the random information. Note that this directly relates to our \cref{eq:discrete-description}, using $\p(h) = 2^{-|h|}$, with the only difference that the optimization is done over programs instead of English sentences. One of the main contributions of our work is to show that something akin to the structure function can be used to define a notion of conceptual distance, and to introduce a more computationally feasible stochastic relaxation of the framework. However, in its pure form this approach still fails. It can be shown that  algorithmic minimal sufficient statistics are trivial for most data \cite[Corollary III.13]{gacs2001algorithmic} and, when non trivial, they do not capture useful semantic information but rather capture the mutual information between the sample and the halting problem \cite[Theorem III.24]{gacs2001algorithmic}.
As we show later, this is a symptom of a more general problem where no expressive enough class of descriptions (such as general programs) can define non-trivial semantic information. This motivates our choice to restrict descriptions to (non-canonical) subclasses such as language sentences

\begin{table*}[t]
    \centering
    \resizebox{.85\textwidth}{!}{
    \begin{tabular}{llcccccccc}\toprule
         & Zero-Shot Method & STS-B & STS12 & STS13 & STS14 & STS15 & STS16 & SICK-R & Avg \\
        \midrule
        \multirow{2}{*}{Paragon} & CLIP-ViTL14 \cite{radford2021learning} & 65.5 & 67.7 & 68.5 & 58.0 & 67.1 & 73.6 & 68.6 & 67.0 \\
        & SimCSE-BERT \cite{gao2021simcse} & \textbf{68.4} & \textbf{82.4} & \textbf{74.4} & \textbf{80.9} &\textbf{78.6} & \textbf{76.9} & \textbf{72.2} & \textbf{76.3} \\
        \midrule
        \multirow{4}{*}{LLaMA \cite{touvron2023llama}} & Cond. Likelihood & 44.3 & 20.8 & 51.8 & 38.6 & 56.0 & 50.9 & 56.7 & 45.6 \\ 
        & Meaning as Trajectories \cite{liu2023meaning} & 70.6 & 52.5 & 65.9 & 53.2 & 67.8 & \textbf{74.1} & 73.0 & 65.3 \\
        & Ours & \textbf{72.3} & \textbf{56.7} & \textbf{67.9} & \textbf{56.9} &\textbf{68.7} & \textbf{74.1} & \textbf{74.3} & \textbf{67.3} \\
        \bottomrule
    \end{tabular}
    }
    \caption{\textbf{Text-to-Text semantic similarity benchmarks.} We evaluate our method using Spearman correlation ($\times 100$) on the STS sentence similarity benchmark. Our method outperforms all zero-shot methods based on vector embeddings. On the SICK-R dataset involving compositional knowledge, conceptual similarity outperforms even the contrastive-trained paragon by $2.1\%$. This shows that our definition of Conceptual Distance is better aligned with human judgement than standard vector embedding. It also outperforms \cite{liu2023meaning}, which can be seen as a particular case of our method for a fixed value of $\lambda=1$, while greatly improving interpretability. This shows the importance of considering different levels of capacity to recover human judgement.
    }
    \label{tab:text-benchmarks}
    \vspace{-0.5em}
\end{table*}

\paragraph{Conceptual Distance.} Our method attempts to tackle all the problems (i)-(iii). First, it does not focus on discriminating samples. Rather we independently find the descriptions of each sample, and then evaluate them on each other. This prevents the method from adversarially picking features that may be good discriminators, but that are not good descriptors -- thus addressing (iii). However, we need to ensure that the description will focus on structural properties of the image, and not on random details. Similarly to Kolmogorov's Structure Function approach, we accomplish this by finding optimal descriptions under a capacity constraint which naturally leads the distance to ignore differences due to incompressible randomness for small values of $C$.

Aside from increasing robustness to unimportant differences, focusing on the initial part of the curve as $C$ grows has other advantages. We posit that human perception of semantic similarity relates to how quickly the distance function grows, rather than its asymptotic value. This motivates our choice of using the AUC of $d_{x_1,x_2}(C)$ up to some small value of $C$ to measure similarity. Indeed, in our experiments we observe that the AUC computed up to some relatively small value of $C$ is better aligned with human similarity annotations than the exact value at any specific $C$, and that using larger values of $C$ decreases alignment.

Our distance implicitly defines a separation between structural and random components of the images. This depends on their ``compressibility'' and is in turn dependent on the choice of hypothesis space, encoder and decoder. Our choice of using language as the hypothesis space, and a particular conditional generative model, is not canonical prompting the question of whether a more canonical choice, like Kolmogorov's choice of using generic programs, would be better. However, as the following theorem shows, any model class which is too expressive would not be able to recover a meaningful notion of structural information.

\paragraph{Theorem.} (There are no canonical definitions of structure, informal). Let $H$ be a class of hypotheses and let $p(x|h)$ be the corresponding decoder. If the decoder $p(x|h)$ is expressive enough to perform perfect test-time optimization, then all samples have the same structure, and the conceptual distance between any pair of samples is zero.

The lack of useful structure emerging from general program classes motivated us to consider an appropriate sub-class that could lead to a more useful, though necessarily non-canonical as expressed in property (ii), measure of conceptual similarity. 
Copyright doctrine acknowledges the absence of a computable notion of similarity, and instead relies on judges and juries to adjudicate each case.  We analogize this human decision-making process to a computable machine-learning process, in which \emph{natural language programs} attempt to describe and distinguish the works in question. 
To operationalize this notion, we reframed Kolmogorov complexity in \cref{sec:concept} to use natural language as opposed to formal language programs for encoding inputs. Natural language descriptions are generated by models that aggregate subjective assessments not just of a single judge or a small group of jurors, but from textual expressions of the millions of individual contributions used to train large-scale models. While our approach makes a subjective choice of using natural language as a representation of the data, such a representation is naturally fit to the task, since natural language is routinely used, and arguably has evolved, to express and communicate abstract concepts.

\section{Practical distance computation}
\label{sec:distance-computation}

To compute our conceptual distance we need to solve the constrained optimization problem \cref{eq:stochastic-description}. The solution can easily be written using the Lagrange multiplier method (see Appendix) and leads to the family of optimal distributions:
\begin{align}
    q^*_x(h|C) &= \frac{1}{Z_\lambda} \p(h) p(x|h)^\lambda,
\end{align}
where $Z_\lambda = \E_{\p(h)} [p(x|h)^\lambda]$  is the normalization factor and $\lambda = \lambda(C) \geq 0$ has to satisfy the condition:
\begin{align}
    \KL{q^*_x(h|C)}{p_\text{code}(h)} = C.
\end{align}
Having $q^*_x(h|C)$ we can compute the various quantities in the definition of the distance. However, given that $q^*_x(h|C)$ is a distribution over an infinite hypothesis space $H$, we cannot represent it explicitly. Fortunately, all the quantities involved are in the form $\E_q[f(h)]$ for some function $f(h)$ and can be estimated through importance sampling. Let $\pi(h)$ be a proposal distribution and let $h_1, \ldots, h_N \sim \pi(h)$ be samples from it. Then we have (see Appendix):
\begin{align*}
    \E_{q^*}[f(h)] &=\textstyle  \E_{\pi}\Big[\frac{q^*(h)}{\pi(h)} f(h)\Big]
    \approx \sum_{i=1}^N \alpha_{\lambda,i} f(h_i)\\
    \alpha_{\lambda, i} &:= \textstyle \operatorname{softmax}\Big(-\lambda \ell(x|h_j) + \log \frac{\p(h_j)}{\pi(h_j)}\Big)_i
\end{align*}
Hence, computing the importance weights $\alpha_{\lambda, i}$ for each sample $h_i$ (all needed quantities are available) we can easily compute an unbiased estimator of our distance function.

\paragraph{Using only the encoder model}
Typically, we have access to an encoder model $p(h|x)$, which outputs description $h$ given the data $x$ --- {\em e.g.,} a captioning model --- but may not have a corresponding generative model $p(x|h)$ to evaluate the likelihood of the data given the description. This problem can be circumvented. By Bayes's rule we have \[p(x|h) = \frac{p(h|x)p(x)}{p(h)}.\]
The encoder model directly provides $p(h|x)$. The likelihood $p(x)$ of the data is however is difficult to evaluate for high-dimensional data. Luckily, the role of $p(x)$ simplifies in our definition of distance (see Appendix) so the term is not needed. Finally, $p(h)$ --- which should not be confused with $\p(h)$ --- is the marginal $p(h) = \sum_{x \in X} p(h|x) p(x)$, which can be estimated through sampling. In practice however, we found that the simple approximation:
$p(h) \approx \frac{1}{2} p(h|x_1) + \frac{1}{2} p(h|x_2)$
performs well and can be computed for free using only the available two samples.

\paragraph{Choice of $\pi(h)$.} To reduce the variance of importance sampling, the proposal distribution $\pi(h)$ should be well aligned with the distributions we want to compute the expectation with, in our case $q_1(h|C)$ and $q_2(h|C)$. Recall that
\[
q_i(h|C) = \frac{1}{Z_\lambda} \p(h) \Big(\frac{p(h|x)}{p(h)}\Big)^{\lambda(C)}
\]
which for $\lambda(C)=1$ simplifies to the posterior $p(h|x_i)$, and will generally be close to it at most important values of $\lambda$, making it a good choice for the proposal distribution. Since we want to be able to use the same samples for both $q_1(h|C)$ and $q_2(h|C)$, we define the proposal distribution as:
\[
\pi(h) := \frac{1}{2} p(h|x_1) + \frac{1}{2} p(h|x_2).
\]
Sampling from this distribution can be achieved easily by sampling $\frac{N}{2} + \frac{N}{2}$ samples from $p(h|x_1)$ and $p(h|x_2)$. 

\paragraph{Choice of {\rm $\p(h)$}.} The choice of the encoding distribution $\p(h)$ can be used to align the conceptual distance with human similarity assessments. A natural choice is to take $\p(h)$ to be the likelihood assigned to the description $h$ by a LLM. Another choice, which is more natural from a computational perspective is to take $\p(h) = \pi(h)$ in order to reduce the variance of the importance sampling estimation. We found the first choice to give more interpretable results (and we use it in our qualitative plots), whereas the latter choice performs better on large scale benchmarks.

\section{Experiments}\label{sec:experiments}

The goal of this section is to evaluate how well our notion of conceptual distance correlates with ground-truth human annotators. 
To this end, we empirically compare samples under three different domains --- text-to-text, image-to-image, and text-to-image (cross-modal) similarity.

\paragraph{Model choice.} %
To implement our proposed similarity score, we require a model $\p(h)$ to compute code length of text descriptions $h$, as well as conditional distributions like $p(h|x)$, where $x$ can represent text or images.  For text encoding we use $\p(h) = \frac{1}{2}\big(p(h|x_1) + p(h|x_2)\big)$ on benchmarks, while on qualitative experiments we use the likelihood of $h$ outputted by LLaMA-13B~\cite{touvron2023llama}. We also use LLaMA-13B as conditional text encoder $p(h|x)$, while when $x$ is an image we use LLaVA \cite{llava}. In the latter case, images are encoded as a sequence of tokens using a vision transformer, and then fed into a general sequence transformer that is trained on both image and text tokens. 

\paragraph{Importance sampling.} For comparing a pair of samples $(x_1, x_2)$, we use descriptions $H_\text{sample} = H_1 \cup H_2$ sampled from the proposal distribution $\pi(h)$, where $H_1 \sim p(h|x_1)$ and $H_2 \sim p(h|x_2)$ are the set of descriptions sampled from $x_1$ and $x_2$ respectively. To ensure fair comparison, we use the same sampling procedure and hyperparameter choices on \cite{liu2023meaning} and  generate from each input $20$ descriptions of $20$ tokens through multinomial sampling.

\paragraph{Sentence similarity.} 
First, we establish whether CC:DAE is able to compute a meaningful notion of conceptual similarity between sentences which aligns with ground-truth human annotations.
Towards this end, we leverage the Semantic Textual Similarity (STS) benchmark \cite{agirre2012semeval,agirre2013sem,agirre2014semeval,agirre2015semeval,agirre2016semeval,cer2017semeval,marelli2014semeval}, where each sentence pair ($x_1, x_2$) is labelled with a similarity score attributed by human annotators. To quantify this alignment, we measure the Spearman correlation between our conceptual distance and the ground-truth scores.
In particular, for each input sentence $x$, we sample continuations $h \sim \pi(h) = \frac{1}{2}(p(h|x_1) + p(h|x_2))$ of the two sentences, which we use as descriptions to estimate their distance.
Our results in \cref{tab:text-benchmarks}  demonstrates that our method correlates well with human judgements. In particular, conceptual similarity outperforms all methods in the literature that have not been explicitly trained on human annotations, with a relative decrease in average error by $-5.8\%$ compared to the next best method, which also uses the same backbone. Moreover, our method outperforms CLIP-ViTL14, which as been trained explicitly on contrastive objectives. Lastly, on the SICK-R (Sentences Involving Compositional Knowledge) dataset, conceptual similarity achieves state-of-the-art performance among zero-shot methods, showing that our method excels at representing compositional structures present in the data.

\begin{table}[t]
    \centering{
    \footnotesize
    \begin{tabular}{cccc}\toprule
        Method & CxC-SIS & CxC-SITS & Average \\
        \midrule
        CLIP-ViTB/16 & 72.08 & 64.60 & 68.34 \\
        Cond. Likelihood & - & 29.46 & - \\
        Meaning as Trajectories \cite{liu2023meaning} & 81.47 & 67.63 & 74.55 \\
        \midrule
        Ours & \textbf{81.44} & \textbf{67.71} & \textbf{74.58}  \\
        \bottomrule
    \end{tabular}
    }
    \caption{\textbf{Image-Image/Image-Text Semantic similarity benchmarks.} Evaluation on human-alignment of similarity scores on the CxC Semantic Image Similarity (SIS) and Semantic Image-Text Similarity (SITS) benchmarks. CC:DAE outperforms CLIP, which is trained explicitly on the multi-modal similarity task, showing that CC:DAE better captures human intuition compared to contrastive methods. On CxC-SITS the method outperforms baselines using the same backbone and, by leveraging the alignment prompt used by \cite{liu2023meaning}, our method achieves comparable performance while benefiting from enhanced interpretability.}
    \label{tab:cxc-benchmarks}
    \vspace{-0.8em}
\end{table}

\begin{figure*}[!t]
    \centering
    \includegraphics[width=.95\linewidth]{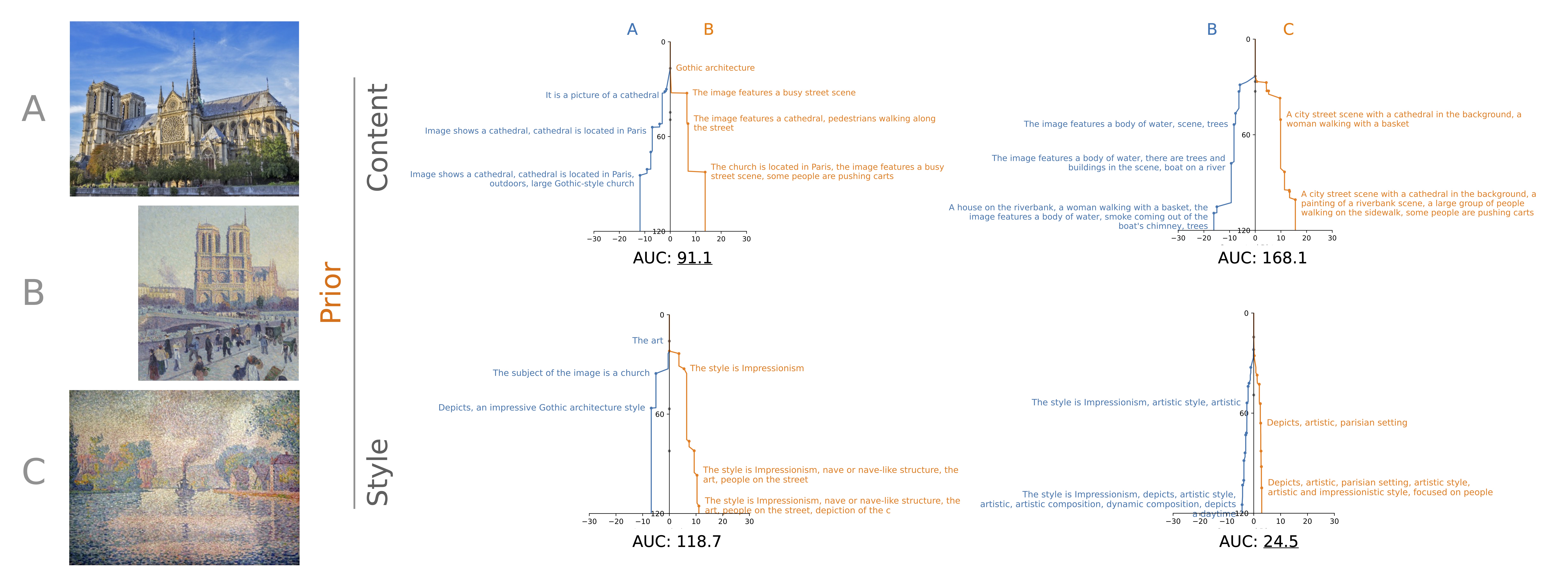}
    \caption{\textbf{Role of prompts.} Consider the three images above: which pair is most similar? This depends if we focus on content --- the first two depict Notre-Dame, the third a boat on the Seine --- or on the style/artistic technique --- the first is a photograph, the second and third are paintings in the pointillist style. By changing the prompt (``Describe the style of the image'' or ``Describe the content of the image''), the user can bias $p(h|x)$ to focus the conceptual distance on one or the other aspect. Note that images A and B are closer under the content prompt, but B and C are closer under the style prompt.
    }
    \label{fig:prior}
    \vspace{-0.5em}
\end{figure*}

\paragraph{Image similarity.} Next, we want to determine whether our conceptual distances correlates with that of human annotators when used to compare images. To accomplish this, we leverage the Crisscrossed Captions Semantic Image Similarity (CxC-SIS) \cite{parekh2020crisscrossed} benchmark containing pairs of images annotated by humans with similarity scores ranging from 0 to 5. To evaluate our method, we measure the Spearman correlation of conceptual distance and human annotations. In \cref{tab:cxc-benchmarks}, we show that our method outperforms all prior zero-shot methods on CxC-SIS, including reducing the error $-33.5\%$ 
relative to CLIP. This supports the idea that our method, based on comparing samples via descriptions, fares better than methods based on discrimination.

\paragraph{Cross-modal similarity.} 
We further investigate whether conceptual similarity can remain applicable when inputs are not assumed to share a common modality. In particular, we extend our comparisons to the CxC-SITS Semantic Image-Text Similarity (CxC-SITS) benchmark, which provide human annotated scores on (image, text) pairs based on their semantic similarity. Our experiments in \cref{tab:cxc-benchmarks} demonstrate that our method indeed generalizes well to cross-modal comparisons, reducing error relative to CLIP by $-8.8\%$.
To further judge the improvement of our method over the base capability of the backbone, on CxC-SITS we compare against a baseline that selects captions by directly comparing the conditional likelihood of the text given the image. We note that this baseline fares poorly, which is aligned with results showing that perplexity is be a poor metric for evaluating alignment of captions \cite{wang2022perplexity}. CC:DAE bypasses this limitation by autoencoding both text and images in a common text description space. CC:DAE also matches or outperform --- both in \cref{tab:text-benchmarks,tab:cxc-benchmarks} ---  the best zero-shot method \cite{liu2023meaning}, which can be seen as a very particular case of our method (see Appendix) while it benefits from enhanced interpretability.

\paragraph{Qualitative results.} In \cref{fig:splash-figure1,fig:prior} we illustrate the behavior of our conceptual distance. To generate richer and more interpretable samples from the description space --- and to further highlight the trade-off between coding length $C$ and distance --- rather than sampling directly $h \sim p(h|x_i)$ we use a beam search to generate effective descriptions of increasing length (see Appendix). For visualization purposes, rather than optimizing a distribution, we restrict the optimization to select only the single best performing description among the samples for each $C$ and display it. In all cases, we see that as expected the distance is zero when restricted to short description (small $C$). As descriptions become more detailed, the distance starts increasing. Each increase in distance can be interpreted looking at the selected $h$. While selected descriptions highlight the differences, we can also look at the best description $h$ explaining both images at the same time (see Appendix) to understand what common structure the two images contain which justifies why the distance is not larger.

\paragraph{Prompting.} Conceptual distance is influenced by the choice of the description space $H$ and the decoder $p(x|h)$, or equivalently the encoder $p(h|x)$. In \cref{sec:motivations} we proved that making such non-canonical choices is inevitable. In \cref{fig:prior} we show that freedom to modify the encoder probability --- for example through prompting --- is indeed not a disadvantage but a helpful option. Different prompts such as ``Describe the style of the image'' or ``Describe the content of the image'' allow the user to easily bias the conceptual distance to focus on the similarity axis of interest.

\vspace{-.2cm}

\section{Conclusions}

We introduce CC:DAE, a principled notion of conceptual similarity which aims to explore and resolve past issues in the definition of semantic distance between high dimensional data. Emprically, the method achieves state-of-the-art results for matching human similarity judgments without fine-tuning on human scores. We built on on Kolmogorov’s central insight that structured information should be easier to explain than randomness. While using general programs as explanations does not yield acceptable results, selecting natural language sequences as our ``programs’’ is enough to spotlight the properties in images and text that are relevant to humans. 
Our general strategy can flexibly incorporate other types of explanations, including visual features, to specify aspects of conceptual similarity that are relevant for specific contexts. In our current implementation, the description space is limited to textual description. This prevents the method from efficiently describing similarities deriving from similar visual arrangement, which is however an important space. We leave to future works extension of the method using more generic decoders that can take both textual and visual features in input.

{
    \small
    \bibliographystyle{ieeenat_fullname}
    \bibliography{main}
}

\clearpage

\input{appendix}

\end{document}

%% file: preamble.tex
\usepackage{enumitem}
\usepackage[dvipsnames]{xcolor}

%% file: new_commands.tex
\newcommand{\p}{p_\text{code}}

%% file: appendix.tex
\appendix

\begin{strip}
    \centering
    \includegraphics[width=0.9\linewidth]{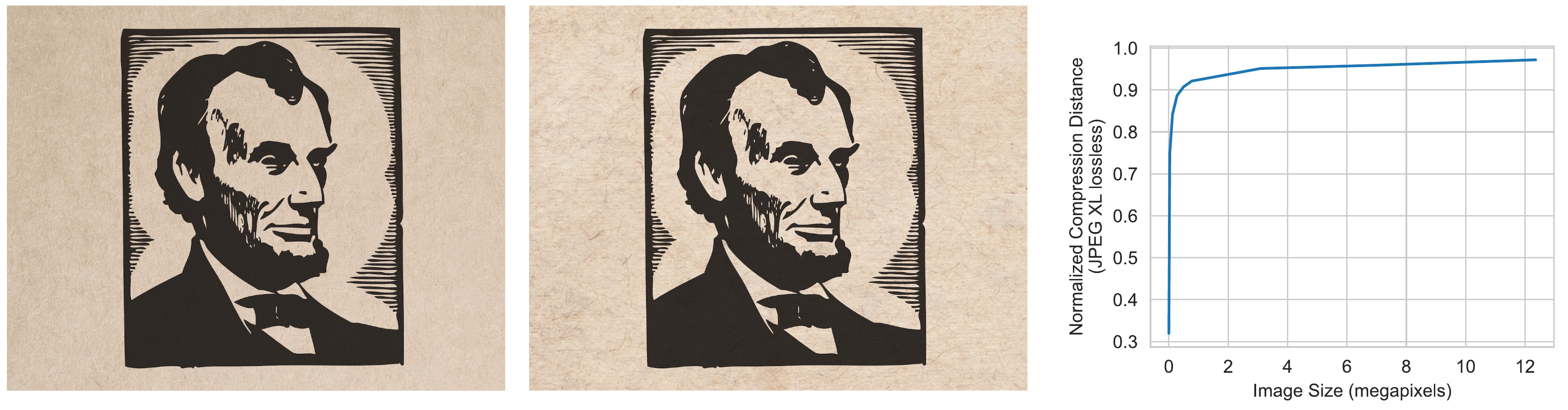}
    \captionof{figure}{
    \textbf{Conceptual Similarity is not the same as shared information.} {Are these two pictures similar?} Not according to Normalized Compression Distance, which measures their difference at 97.1\% (estimated using JPEG XL lossless compression). However, they share all the structural information --- they are the exact same ink print on a piece of paper. The only difference is the randomness of the paper texture. Most people would not consider it a significant conceptual difference, but since NCD cannot differentiate structure from randomness, this slight change accounts for 97.1\% of the difference. This problem is inherent in high-dimensional data where information in random variation overshadows structural information. In fact, on the right we plot the NCD distance as we change the resolution of the image, showing that the distance increases drastically as we increase the dimension of the data.}
    \label{fig:ncd-distance}
\end{strip}

\begin{table*}
  \centering
  \small
  \scalebox{0.8}{
    \begin{tabular}{ll ccccccc} 
    \toprule
     && \multicolumn{3}{c}{\bf REPLACE}  &  \multicolumn{2}{c}{\bf SWAP} & \multicolumn{2}{c}{\bf ADD} \\ 
     \cmidrule(lr){3-5}\cmidrule(lr){6-7}\cmidrule(lr){8-9}
     
    \textbf{Source} &\textbf{Model} & \textbf{Object} & \textbf{Attribute} & \textbf{Relation} & \textbf{Object} & \textbf{Attribute} & \textbf{Object} & \textbf{Attribute}   \\ 
    \midrule

\multirow{1}{*}{OpenAI~\cite{radford2021learning}}    
& RN50x64&94.5& 83.5& 70.6& 61.8& 66.7& 83.3& 74.0\\
\midrule
\multirow{2}{*}{LAION~\cite{schuhmann2022laion}} 
& ViT-bigG-14& 96.7& 88.1 & 74.8& 62.2& 74.9& 92.2& 84.5\\ %
& xlm-roberta-large-ViT-H-14& 96.9 & 86.0& 72.1& 63.8& 72.1& 93.1 & 86.1\\
\midrule
\multirow{1}{*}{DataComp~\cite{gadre2023datacomp}}   
& \texttt{xlarge:}ViT-L-14& 95.5& 84.5& 67.0& 65.0& 66.8& 91.0& 85.0\\
\midrule
\multirow{3}{*}{LLaVA\cite{llava}} 
& Cond. Likelihood & 78.8 & 77.7 & 73.3 & 77.6 & 86.0 & 36.2 & 76.2 \\
& Meanings as Trajectories \cite{liu2023meaning} & 90.4 & 80.6 & 78.8 & 69.9 & 76.6 & 75.7 & 82.8 \\
& CC:DAE (ours) & 91.0 & 82.1 & 82.2 & 73.6	& 78.8 & 77.0 & 86.0 \\
\bottomrule
\end{tabular}
}
\caption{
\textbf{Performance on SugarCrepe multi-modal image-caption alignment benchmark.} We show that CC:DAE can be extended to compute similarity between data of different modalities. CC:DAE outperforms or matches all the baseline contrastive-based models (numbers from \cite{hsieh2023sugarcrepe}) on 4 out of 7 tasks. Compared to methods using our same backbone, we significantly outperform the conditional likelihood baseline. We also uniformly outperform \cite{liu2023meaning} which uses the same backbones and trajectories as our method.} 
\label{tab:sugarcrepe}
\end{table*}

\section{Limitations of Information Theoretic Distances}

In \cref{fig:ncd-distance} we show that two images that are conceptually identical are considered almost completely different by the Normalized Compression Distance (NCD), an information theoretic distance. As we now discuss, this is an intrinsic problem of information theoretic distance, which cannot distinguish differences due to structural properties from differences due to randomness, the latter becoming dominant as the number of dimensions grows (\cref{fig:ncd-distance}, right).

NCD \cite{li2008introduction} defines the distance between samples in terms of their common (algorithmic) information. Let $x$ and $y$ be two samples. We denote with the Kolmogorov complexity $K(x)$ the length of the shortest program that can output $x$ (equivalently, up to a constant, its best possible compression cost using  commutable function), and with $K(x|y) = K(xy) - K(y)$ the length of the shortest program that can reconstruct $x$ given $y$ as input. If $x$ does not contain any information that is not already contained in $y$, then $K(x|y) \approx 0$ and we can consider $x$ to be similar to $y$. Making the role of $x$ and $y$ symmetric, this motivates the following definition:
\begin{align*}
\operatorname{NCD}(x, y) &= \frac{\max\{K(x|y), K(y|x)\}}{\max\{K(x), K(y)\}} \\
&= \frac{K(xy) - \min\{K(x), K(y)\}}{\max\{K(x), K(y)\}}.
\end{align*}
While this is a theoretically viable definition, in practice the shortest coding length $K(x)$ cannot be computed explicitly. However, the distance can be approximated using a strong compression algorithm $Z(x)$ as follows:
\begin{align*}
    \operatorname{NCD}_Z(x, y) &= \frac{Z(xy) - \min\{Z(x), Z(y)\}}{\max\{Z(x), Z(y)\}}.
\end{align*}
On the surface, this distance is well positioned to capture shared algorithmic structure, and hence recover meaningful similarities between the samples. However, this is not the case when data can be noisy. The following proposition shows that two pictures that differ only by some slight noise --- for example two consecutive photos differing only by sensor noise ---  always have close-to-maximal NCD.

\paragraph{Proposition.} Let $s$ be an image 
 of low complexity $K(s)$. Suppose that two measurements $x := s \oplus n_x$ and $y := s \oplus n_y$ are generated by adding Bernoulli noise $n_x, n_y \sim \operatorname{Bern}(p)$ to $s$, where $\oplus$ denotes the bit-wise XOR. Denote with $H(p)$ the entropy of the Bernoulli distribution. Then, in the limit of large dimension $D = |s|$ we have:
 \[
 \operatorname{NCD}(x,y) \approx 1 - \frac{K(s)}{D\, H(p)} \xrightarrow[]{D \to \infty} 1
 \]
 
\begin{proof}
If the noise $p$ is small, the optimal way to compress $x$  --- and similarly $y$ --- is to simply encode both $s$ and $n_x$ independently:
\[
K(x) \approx K(s) + K(n_x) = K(s) + |s| H(p).
\]

Similarly, the cost of encoding $x$ and $y$ together is the cost to encode (once) the shared $s$ and the two noise masks:
\[
K(xy) = K(s) + K(n_x) + K(n_y) = K(s) + 2 |s| H(p).
\]
Putting all together we get:
\begin{align*}
\operatorname{NCD}(x, y) &= \frac{|s| H(p)}{K(s) + |s| H(p)} 
=  \frac{1}{1 + \frac{K(s)}{|s| H(p)}} \\
&= 1 - \frac{K(s)}{|s|H(p)} + o\Big(\frac{K(s)}{|s|H(p)}\Big)
\end{align*}
as we wanted.
\end{proof}

To demonstrate this effect empirically, in \cref{fig:ncd-distance} we generate two images $x$ and $y$ using the same basic picture $s$, but adding two different noise pattern $n_x$ and $n_y$ (the different paper textures). We then compute $\operatorname{NCD}_Z$ using as compressor $Z$ the recent JPEG XL lossless codec (which gave the best compression across the tried codecs). Since $Z$ does not support compression of two images simultaneously, instead of $Z(xy)$ we use the lower-bound:
\[
Z(xy) \geq Z(x) + Z(y) - Z(s).
\]
The empirical behavior --- even if using a suboptimal compression scheme and using correlated noise --- indeed follows the theoretical prediction.

\section{Multi-Modal Similarity on SugarCrepe}
Our method can also be used to compute similarity between data in different modalities, as long as they share the description space $H$. To test this, we evaluate our method on SugarCrepe \cite{hsieh2023sugarcrepe}, a vision-language compositionality benchmark framed as a binary classification task: given an image and a pair of candidate captions, the task is to select the right caption for the image. Negative captions in each pair are generated from the ground-truth caption as ``compositional distractors" via replacing, swapping, or adding atomic concepts. Since caption pairs differ only by an atomic concept, effective methods require capturing compositional structures in both image and text modalities. We use using CC:DAE for classification by computing the conceptual distance between the image and each caption, and selecting the caption that yields the lowest distance. 

For our experiments on SugarCrepe, we generate via multinomial sampling 10 trajectories of maximum 10 tokens from each input image or candidate caption as descriptions for our method. For fair comparison, we apply the same settings for the Meaning as Trajectories baseline. 

In \cref{tab:sugarcrepe}, we show that conceptual similarity matches or outperforms all multi-modal contrastive-based models (which are trained specifically for this task) on 4 out of 7 tasks. This holds especially for the hardest benchmarks measured by the poor performance of the paragon contrastive models: \textit{Replace Relation}, \textit{Swap Object}, \textit{Swap Attribute}, and \textit{Add Attribute}. Our model also significantly outperforms the conditional likelihood baseline in most tasks --- which notably performs even worse than random guessing on the \textit{Add Object} benchmark --- and also uniformly outperforms \cite{liu2023meaning} on all 7 tasks when using the same sampled descriptions for both methods.

\section{Non-Existence of a Canonical Structure}

In this section we want to prove the following statement.

\paragraph{Theorem.} (There are no canonical definitions of structure, informal). Let $H$ be a class of hypotheses and let $p(x|h)$ be the corresponding decoder. If the decoder $p(x|h)$ is expressive enough to perform perfect test-time optimization, then all samples have the same structure, and the conceptual distance between any pair of samples is zero.

We provide a general sketch of proof in the general setting, and we refer the reader to \cite{gacs2001algorithmic} for a more technical proof in the specific case of $H = \{\text{computable distributions}\}$. First, we need to introduce some additional definitions --- adapted from \cite{vitanyi} ---  to formalize the notion of structure of $x$ as all the non-random information contained in $x$. Given an hypothesis class $H$, consider the function:
\begin{align}
    \label{eq:lambda-function}
    \lambda_x(C) = \min_{h \in H}& \ \ -\log p_h(x) - \log \p(h) \\
    \text{s.t.}& \ \ - \log \p(h) \leq C. \nonumber
\end{align}
This function closely relates to our optimization problem \cref{eq:discrete-description}, and can be seen as the compression code for $x$ using a two part code that first specifies a description $h \in H$ --- with cost $\ell(h) = -\log \p(h)$ --- and then uses it to encode $x$ using $-\log p_h(x)$ bits. As $C$ grows, and we can use better  fitting descriptions, $\lambda_x(C)$ decreases until it reaches a minimum non-zero value, which is the best compression cost achievable using this class. We call a hypothesis $h$ \textit{sufficient statistic} if it witnesses this minimum. A sufficient statistic of $x$ describes all the properties that are compressible under $H$, however it may also encode random bits of incompressible (non-structural) information. To prevent this, we define a \textit{minimal sufficient statistic} as any sufficient statistic $h$ which has minimal coding length $\ell(h)$ (equivalently, $h$ witnesses the first point where $\lambda_x(C)$ reaches its minimum). Intuitively, a minimal sufficient statistic captures all structural properties and no random properties. This definition was proposed by Kolmogorov, and further formalized by \cite{vitanyi} to separate structural and random properties of the data.

We now want to show that, in general, a minimal sufficient statistic will always be trivial if the class $H$ and the decoder function are expressive enough. Hence, picking a canonical class of hypotheses/functions (e.g., all possible functions, all computable functions) always lead to trivial structure and restriction to a smaller non-canonical set (all linear functions, descriptions that are English sentences, etc.) is necessary to talk about structural properties.

\begin{proof} 
Let $H$ be an hypothesis class. Consider the function 
\begin{align}
    \label{eq:lambda-function}
    \lambda_x(C) = \min_{h \in H}& \ \ -\log p_h(x) - \log \p(h) \\
    \text{s.t.}& \ \ - \log \p(h) \leq C, \nonumber
\end{align}
The best compression we can achieve using $H$ is given by the minimum over:
\[
\min_{h\in H} - \log p_h(x) - \log \p(h).
\]
Suppose however that, because the model class is rich enough, there is an hypothesis $h_\text{search}$ which can compute:
\[
\log p_{h_\text{search}}(x) = \min_{h\in H} - \log p_h(x) - \log \p(h).
\]
This can be implemented, for example, by enumerating all elements of $H$ by their coding length $-\log \p(h)$ and testing all of them until a minimum is found (only finitely many have to be tested, since $h$ whose coding length is too long cannot be minima). Then $h_\text{search}$ would be a minimal sufficient statistic \textit{at the same} time for all possible samples, hence all samples really have the same structure. Moreover, since all samples would consider $h_\text{search}$ an optimal description, their conceptual distance would be zero.
\end{proof}

This results motivate our non-canonical choice of restricting to $H = \{\text{natural language sentences}\}$ in defining a conceptual distance. We also note that when $H$ is the class of all possible programs, $h_\text{search}$ may not be computable due to the halting problem. The sketch of the proof however remains valid for most samples, and in particular all the ones likely to occur as real-world measurements (see \cite{gacs2001algorithmic} for an extended discussion).

\section{Connection with Liu et al.~{\cite{liu2023meaning}}}

Let $x_1$ and $x_2$ be two sentences, and let $p(h|x_i)$ be the distribution over the trajectories $h$ that can extend $x_i$, where $p(h|x_i)$ is computed using a large language model. \cite{liu2023meaning} proposes to measures the similarity between $x_1$ and $x_2$ based on the similarity between the distribution of trajectories they can generate. Specifically,
\[
d_\text{traj}(x_1, x_2) = \E_{h \sim \frac{1}{2} (p(h|x_1) + p(h|x_2))} \big| \log p(h|x_1) - \log p(h|x_2) \big|
\]
Interestingly, while this perspective is very different from our definition of conceptual distance, we note that it can be derived as a particular case of our method. In particular, consider using the set $H$ of trajectories as the set of sampled descriptions used to compute the conceptual distance. When using the encoder-only method to compute our distance we have:
\begin{align}
    q^*_{x_i}(h|C) &= \frac{1}{Z_\lambda} \p(h) \Big(\frac{p(h|x_i)}{p(h)} \Big)^\lambda,
\end{align}
In the case of language, we can assume that the marginal probability $p(h) = \int p(h|x) p(x) dx$ of the trajectory $h$ over all  possible prefixes $x$ should be similar to unconditional likelihood $\p(h)$ of the text. With this assumption, and in the particular case of $\lambda(C) = 1$, the optimal distribution of descriptions reduces to:
\begin{align*}
    q^*_{x_i}(h|\lambda=1) &= p(h|x_i).
\end{align*}
Using this distribution to compute the conceptual distance we have:
\begin{align*}
    d_{x_1, x_2}(\lambda=1) =&\, \E_{p(h|x_1)} [\log p(h|x_2) - \log p(h|x_1)] \\
    &\,+\E_{p(h|x_2)} [\log p(h|x_1) - \log p(h|x_2)] 
\end{align*}
assuming that $\log p(h|x_2) - \log p(h|x_1)$ is mostly positive when $h$ is sampled from $p(h|x_1)$ --- and vice versa for $x_2$ --- we can rewrite this with absolute values as:
\begin{align*}
    d_{x_1, x_2}(\lambda=1) &=\, \E_{p(h|x_1)} |\log p(h|x_2) - \log p(h|x_1)| \\
    &\hspace{3em}+\E_{p(h|x_2)} |\log p(h|x_1) - \log p(h|x_2)| \\
    &\hspace{-4em} =\, 2 \cdot \E_{\frac{1}{2} p(h|x_1) + \frac{1}{2} p(h|x_2)} \big| \log p(h|x_1) - \log p(h|x_2) \big|.
\end{align*}
Hence we can see $d_\text{traj}(x_1, x_2)$ as a particular case of our conceptual distance when using a particular capacity $C$ such that $\lambda(C) = 1$, and making a particular choice of using trajectories as descriptions of the sentences $x_1$ and $x_2$. Our results in \cref{tab:text-benchmarks,tab:sugarcrepe} show that using our distance without these restrictions outperforms \cite{liu2023meaning} when evaluated in the same setting.

\section{Using Only the Encoder Model}
As described in \cref{sec:distance-computation}, for image experiments we use the LLaVA image-to-text encoder $p(h|x)$ to evaluate the likelihood $p(x|h)$, through Bayes' rule:
\[
p(x|h) = \frac{p(h|x)}{p(h)}p(x).
\]
We now want to show that the term $p(x)$ does not affect the distance computation, and can be ignored. First note using the above identity, we have that the reconstruction loss is
\[
\ell(x|h) = -\log \frac{p(h|x)}{p(h)} - \log p(x) = \bar{\ell}(x|h) - \log p(x),
\]
where we defined $\bar{\ell}(x|h) = -\log \frac{p(h|x)}{p(h)}$.  Considering the optimization problem in \cref{eq:stochastic-description} to find the optimal description under a capacity constrain:
\begin{align*}
    q^*_x(h|C) = \argmin_{q(h) \in \mathcal{P}(H)}& \ \ \E_{h \sim q(h)} [ \ell(x|h)] \\
    \text{s.t.}& \ \ \KL{q(h)}{p(h)} \leq C, 
\end{align*}
we see that $- \log p(x)$ (which does not depend on $h$) only accounts for an additive constant ($\E_{h \sim q(h)} [ \ell(x|h)] = \E_{h \sim q(h)} [ -\log \frac{p(h|x)}{p(h)}] + \log p(x)$). Hence $q^*_x(h|C)$ does not depend on $p(x)$. Using \cref{eq:d}, the distance is:

\resizebox{1\linewidth}{!}{%
\begin{minipage}{\linewidth}%
\begin{align*}
    d_{x_1,x_2}(C) &= \E_{q_1}[\ell(x_1|h)] + \E_{q_2}[\ell(x_2|h)] - \E_{q_\cap}[\ell(x_1|h) + \ell(x_2|h)] \\
    &= \E_{q_1}[\bar{\ell}(x_1|h)] + \log p(x_1) + \E_{q_2}[\bar{\ell}(x_2|h)] + \log p(x_2) \\
    &\hspace{2.1em}- \E_{q_\cap}[\bar{\ell}(x_1|h) + \bar{\ell}(x_2|h)] - (\log p(x_1) + \log p(x_2)) \\
    &= \E_{q_1}[\bar{\ell}(x_1|h)] + \E_{q_2}[\bar{\ell}(x_2|h)] - \E_{q_\cap}[\bar{\ell}(x_1|h) + \bar{\ell}(x_2|h)],
\end{align*}
\end{minipage}
}
\vspace{0.2em}

\noindent and all quantities in the last expression are independent of the value of $\log p(x_i)$.

\section{Closed-Form Expressions}
We now derive the close form expression for the solution of \cref{eq:stochastic-description}:
\begin{align*}
    q^*_x(h|C) = \argmin_{q(h) \in \mathcal{P}(H)}& \ \ \E_{h \sim q(h)} [ \ell(x|h)] \\
    \text{s.t.}& \ \ \KL{q(h)}{\p(h)} \leq C.
\end{align*}
Writing the Lagrangian corresponding to the constrained optimization problem we have:
\begin{align*}
    \mathcal{L} =&\, \textstyle \E_{h \sim q(h)} [ \ell(x|h)] + \gamma \big(  \KL{q(h)}{\p(h)} - C \big) \\
    &\quad\,+ \alpha \big( \int q(h)dh - 1\big)
\end{align*}
Setting to zero the derivatives of $\mathcal{L}$ with respect to $q(h)$ we have:
\begin{align*}
    \partial_{q(h)} \mathcal{L} &= l(x|h) + \gamma \Big( \log \frac{q(h)}{\p(h)} - 1\Big) + \alpha = 0,
\end{align*}
from which we get:
\begin{align*}
    q(h) &= \p(h) \exp\Big(- \frac{1}{\gamma} l(x|h) +1 - \alpha \Big) \\
    &= \frac{1}{Z} \p(h) \exp\big(-\lambda l(x|h)\big) \\
    &= \frac{1}{Z} \p(h) p(x|h)^\lambda,
\end{align*}
where we defined $\lambda := \frac{1}{\gamma}$ and $Z := \exp(\alpha - 1)$, and we used $l(x|h) = - \log p(x|h)$. Enforcing the constraint that $q(h)$ is a probability distribution and integrates to one (note that $q(h) > 0$ is automatically satisfied), we get:
\[
Z = Z_\lambda = \int \p(h) p(x|h)^\lambda dh.
\]
Enforcing the remaining constraint on $\lambda = \lambda(C)$ allows finding the optimal distribution for the particular capacity $C$. While the solution cannot be written analytically, it can be found easily by binary search over $\lambda$. Doing it is however not necessary for our method: since our method only cares about the function
\[
\beta_x(C) = \E_{q^*(h|C)}[\ell(x|h)]
\]
as $C$ varies, rather than solving the constraint we can simply trace the curve
\begin{align*}
\beta_x(\lambda) &= \E_{q^*(h|\lambda)}[\ell(x|h)] \\
C(\lambda) &= \KL{q^*(h|\lambda)}{\p(h)}
\end{align*}
as $\lambda$ varies in order to reconstruct the function $\beta_x(C)$. In our experiments, we sample $\lambda \in \operatorname{linspace}(0, 100, 200)$ and  linearly interpolate the results to approximate $\beta_x(C)$.

\section{Experimental Details for Qualitative Plots}

To improve interpretability of the plot, and aid better understanding of the key components of the method, we use the following setup for our qualitative plots. We use as coding length $\ell(h) = - \log \p(h)$ the log-likelihood assigned to $h$ by a LLaMA language-model. This  scales with the complexity and length of the sentence $h$. In \cref{eq:stochastic-description} the complexity $C$ of the hypothesis distribution $q(h)$ is given by $\KL{q(h)}{p(h)}$ which measures closeness of the distribution to the prior $\p(h)$. This means that, when $C$ is low, the method will select all sentences, giving higher probability to short ones. Since distribution of sentences are difficult to visualize, we force $q(h)$ to be a Dirac delta, thus making it more similar to \cref{eq:discrete-description} and effectively selecting the single best description with coding length $\ell(h) \leq C$.

To observe the effect of varying $C$, it is helpful to sample descriptions that are increasingly longer (larger $\ell(h)$) and more descriptive (smaller $\ell(x|h)$), as we expect longer descriptions to be preferentially picked as $C$ increases. Unfortunately, directly sampling longer descriptions with LLaVA does not result in good descriptions, as the model starts hallucinating information that is irrelevant for the image and thus does not increase $\ell(x|h)$. To remedy this, we use the following beam search method. First, we prompt the LLaVA model to generate short ``atoms'' of information about the image (using the prompt: ``\texttt{Describe in 10 short bullet points what you see in the image. Do not provide explanations.}''). We compute a total of 40 atoms for each image of the pair, and combine them in a common dictionary. We also sample 40 atoms that are good descriptions of both images at the same time, by sampling tokens from the ensemble $\frac{1}{2}\log p(h_t| h_{<t}, x_1) + \frac{1}{2} \log p(h_t| h_{<t}, x_2)$ of the log-likelihoods conditioned on each image. We then use a beam search to combine the atoms in the dictionary in increasingly longer sentences that are optimal description for each image --- i.e., minimize $\ell(x|h)$. Since evaluating $\ell(x|h)$ with LLaVA at each step of the beam search is expensive, we use the CLIP similarity between $h$ and $x$ as a proxy score. For \cref{fig:prior}, we add to the beam search a penalty to preferentially avoid sentences that do not relate to the prompt (``Describe style'' or ''Describing content'') by subtracting from the CLIP score the CLIP similarity between the sentence and the negative prompt. This procedure is used to sample a qualitatively interesting set $H$ of descriptions which helps providing a better intuitive understanding of the conceptual distance. The distance computation is otherwise unaltered. In the  quantitative results we instead simply sample $H \sim \frac{1}{2} p(h|x_1) + \frac{1}{2} p(h|x_2)$.